\documentclass{article} 
\usepackage{iclr2025_conference,times}

\usepackage{hyperref}
\usepackage{url}
\usepackage{graphicx}

\title{Nepotistically Trained Generative Image Models Collapse}

\author{Matyas Bohacek \\
Stanford University \\
\texttt{maty@stanford.edu} \\
\And
Hany Farid \\
University of California, Berkeley \\
\texttt{hfarid@berkeley.edu} \\
}

\iclrfinalcopy 
\begin{document}

\maketitle

\begin{abstract}
  Trained on massive amounts of human-generated content, AI-generated image synthesis is capable of reproducing semantically coherent images that match the visual appearance of its training data. We show that when retrained on even small amounts of their own creation, these generative-AI models produce highly distorted images. We also show that this distortion extends beyond the text prompts used in retraining, and that once affected, the models struggle to fully heal even after retraining on only real images.
\end{abstract}

\section{Introduction}

From text to image, audio, and video, today's generative-AI systems are trained on large quantities of human-generated content. Most of this content is obtained by scraping a variety of online sources~\citep{zhang2024text, srinivasan2021wit, schuhmann2022laion}. As generative AI becomes more common, it is reasonable to expect that future data scraping will invariably catch generative AI's own creations~\citep{martinez2023towards, martinez2023combining}. We ask what happens when these generative systems are trained on varying combinations of human-generated and AI-generated content.

Despite the rapidly accelerating capabilities of generative AI, evidence suggests that retraining a model on its own creation --- what we call model self-poisoning --- leads to artifacts in the output of the newly trained model. It has been shown, for example, that when retrained on their own output, large language models (LLMs) contain irreversible defects that cause the model to produce gibberish~\citep{shumailov2023curse,shumailov2024ai}. Similarly, on the image generation side, it has been shown~\citep{alemohammad2023self} that when retrained on its own creations, StyleGAN2~\citep{karras2020analyzing} generates images (of faces or digits) with visual and structural defects. Interestingly, the authors found that there was a deleterious effect as the ratio of AI-generated content used to retrain the model ranged from $0.3\%$ to $100\%$.

It has been shown that in addition to GAN-based image generation, diffusion-based text-to-image models are also vulnerable~\citep{xing2024ai}. The authors in~\citep{martinez2023towards, martinez2023combining} showed that in a simplified setting, retraining on one's own creation can lead to image degradation and a loss of diversity. Examining the impact of self-poisoning of ID-DPM~\citep{nichol2021improved}, \citet{hataya2023will} report somewhat conflicting results depending on the task at hand (recognition, captioning, or generation). With respect to generation, the authors report a relatively small impact on image quality but do note a lack of diversity in the retrained models.

Building on these earlier studies, we show that the popular open-source model Stable Diffusion\footnote{\url{https://github.com/Stability-AI/StableDiffusion}} (SD) is highly vulnerable to data self-poisoning. In particular, we show that when iteratively retrained on faces of its own creation, the model --- after an initial small improvement --- quickly collapses, yielding highly distorted and less diverse faces. Somewhat surprisingly, even when the retraining data contains only $3\%$ of self-generated images, this model collapse persists. We also investigate the extent of this model self-poisoning beyond the prompts used for retraining and examine the ability of the poisoned model to heal with further retraining on only real images.


\section{Methods}

\subsection{Images}
\label{subsec:images}

Starting with the FFHQ image dataset containing $70{\small ,}000$ faces of size $1024 \times 1024$ pixels~\citep{karras2019style}, we automatically classified~\citep{serengil2021lightface,rezgui2019carthage} each face based on gender (man/woman), race (asian, black, hispanic, indian, and white), and age (young, middle-aged, old). A total of $900$ images were randomly selected constituting $30$ images from each of $30$ demographics ($2$ [gender] $\times$ $5$ [race] $\times$ $3$ [age]).

\begin{figure}[t]
    \centering
    \begin{tabular}{c@{\hspace{0.1cm}}c@{\hspace{0.1cm}}c@{\hspace{0.1cm}}c@{\hspace{0.1cm}}c}
        \includegraphics[width=0.17\linewidth]{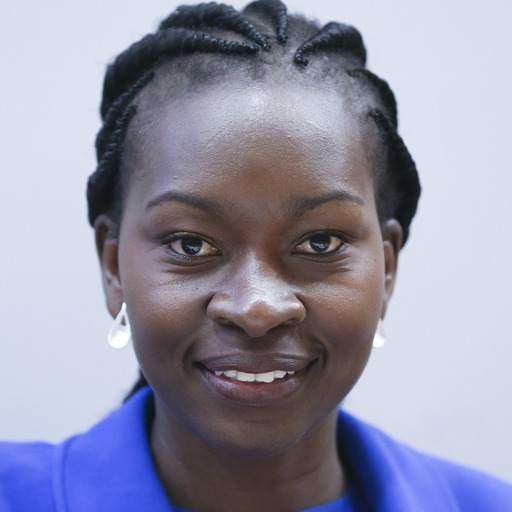} &
        \includegraphics[width=0.17\linewidth]{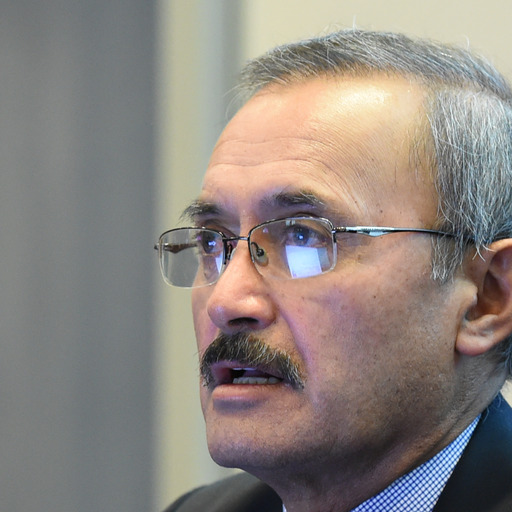} &
        \includegraphics[width=0.17\linewidth]{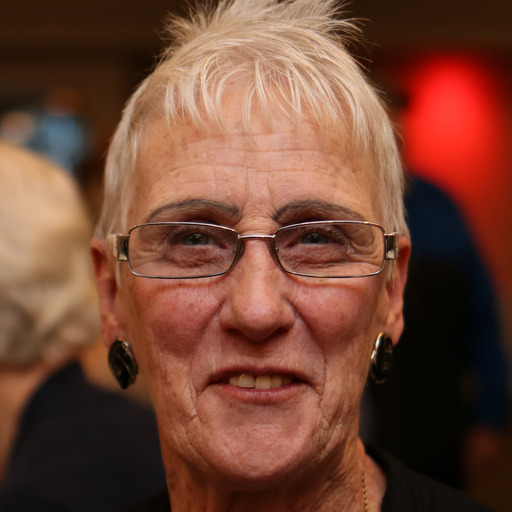} &
        \includegraphics[width=0.17\linewidth]{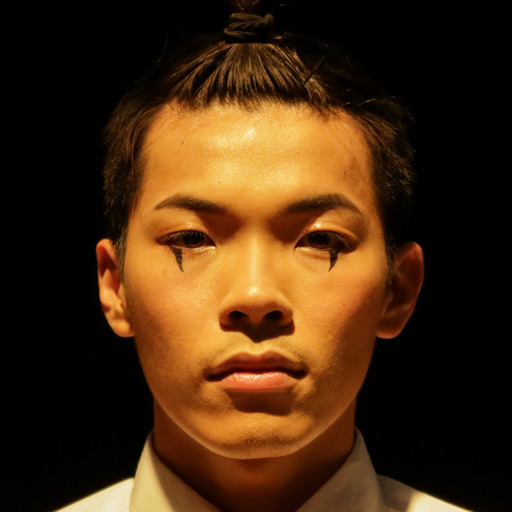} &
        \includegraphics[width=0.17\linewidth]{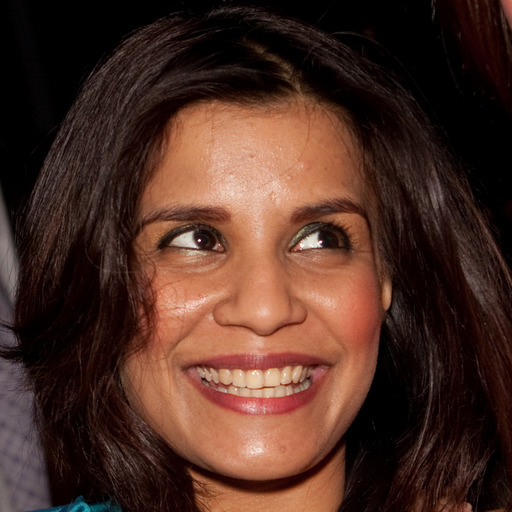} 
        \\
        \includegraphics[width=0.17\linewidth]{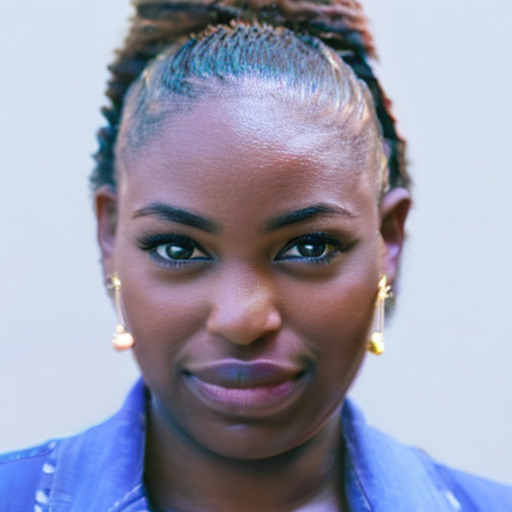} &
        \includegraphics[width=0.17\linewidth]{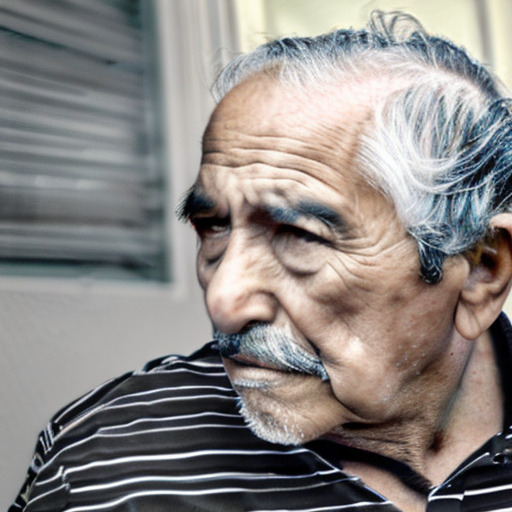} &
        \includegraphics[width=0.17\linewidth]{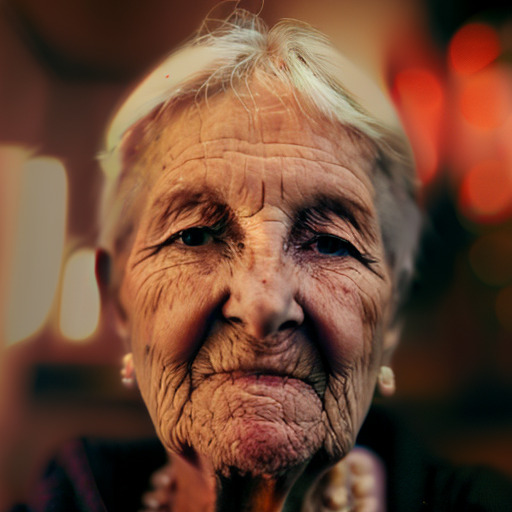} &
        \includegraphics[width=0.17\linewidth]{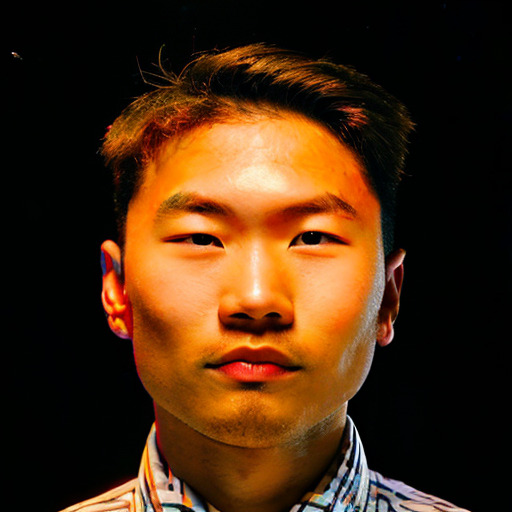} &
        \includegraphics[width=0.17\linewidth]{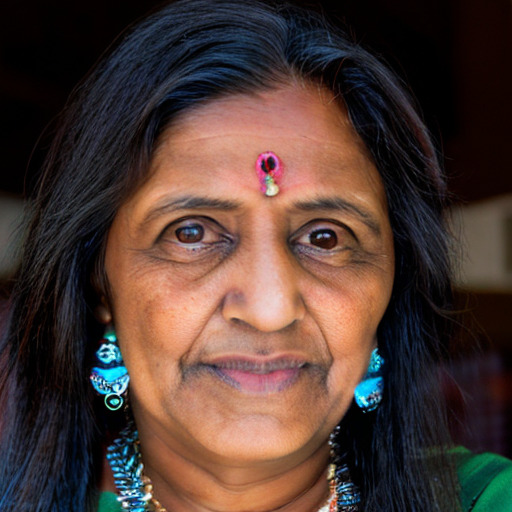}
    \end{tabular}
    \caption{Examples of real images (top) used to seed image-to-image generation (bottom).}
    \label{fig:image-to-image-synthesis}
\end{figure}

These real images were used as input to the image-to-image synthesis pipeline of Stable Diffusion (v.2.1)~\citep{rombach2022high} to generate $900$ images consistent with the demographic prompt ``a photo of a [age] [race] [gender].'' The strength parameter was set to $0.8$ and the number of inference steps to $250$, with all other parameters left at their defaults (a strength of $0.0$ reproduces the input image and a strength of $1.0$ effectively the input image). Shown in Figure~\ref{fig:image-to-image-synthesis} are examples of real (top) and generated faces (bottom). As described next, these $900$ generated faces were used to seed the iterative model retraining. We used this image-to-image synthesis instead of the unconstrained text-to-image generation to balance the comparison of model retraining on healthy data and self-generated data.

%
%

\subsection{Evaluation}
\label{subsec:evaluation}

A standard Fréchet inception distance (FID)~\citep{heusel2017gans} and Contrastive Language-Image Pre-training score (CLIP)~\citep{radford2021learning,hessel2021clipscore} are used to assess the quality of generated images. These metrics are commonly used in text-to-image model benchmarks.

The FID is used to quantify the realism of generated images by comparing the set of generated images to matched real images. This metric compares two sets of $N$ images -- one real and one AI-generated -- with a $1:1$ prompt correspondence. The FID is calculated as the Fréchet distance between the mean and covariance matrices of the corresponding images' Inception-v3 embeddings~\citep{szegedy2015going}. A smaller FID corresponds to a higher similarity between the AI-generated and corresponding real images. In our case, the reference image set consists of $820$ real faces from the FFHQ dataset. This is less than the full set of $900$ because the FFHQ dataset is not sufficiently diverse across all demographic categories. We compare $820$ generated images, consisting of the maximum number of images per $30$ demographic groups, to this set of $820$ demographically similar real images.

The CLIP score calculates the cosine similarity between the image and caption embeddings. This metric relies on the latent space of a pre-trained CLIP model to gauge whether the generated image is semantically consistent with the caption. In our case, the CLIP score is evaluated across all $900$ generated images using the CLIP ViT-H/14 model trained with the LAION-2B English subset of LAION-5B\footnote{\url{https://huggingface.co/laion/CLIP-ViT-H-14-laion2B-s32B-b79K}}. A larger CLIP score corresponds to higher semantic coherence.

\subsection{Poisioning}
\label{subsec:poisioning}

We define model self-poisoning as the process of retraining a generative model on data with varying amounts of images generated by a base model. Unlike adversarial training~\citep{carlini2023poisoning}, the training image captions are consistent with the image contents.

The architecture of Stable Diffusion comprises three main modules: CLIP text encoder, U-Net, and variational autoencoder (VAE). The CLIP text encoder converts the user text prompt into a CLIP embedding, a high-dimensional representation in a multimodal vector space. The base SD model does not train a new CLIP text encoder but instead uses a pre-trained model with frozen weights. The CLIP embedding is provided as input into the U-Net module. The U-Net, guided by a scheduler, gradually denoises what was originally a random noise image towards a representation that is semantically consistent with the text embedding. At the end of this diffusion process, the VAE converts the latent U-Net image representation into pixel space to yield a final image.

The model retraining proceeds as follows. Leaving the CLIP text encoder and VAE modules intact, we retrained the denoising U-Net module of the base Stable Diffusion model (SD v.2.1) using the recommended parameters (a constant learning rate of $2 \times 10^{-6}$, $50$ epochs, $512 \times 512$ image resolution, no mixed precision, and random horizontal flip augmentation). The rationale for this is that these modules handle supportive tasks -- the conversion from text to the latent space and from the latent space back to the pixel space -- outside of the core diffusion process. Freezing of these modules is standard practice for retraining. The model was initially retrained on the $900$ SD-generated images and demographic captions. Another $900$ images with the same demographic prompts were generated using this retrained model. These images were then used to retrain the model. This process was repeated for a total of five iterations.

This entire process was repeated with different compositions of faces ranging from the above $100\%$ SD-generated and $0\%$ real faces to a $50\%/50\%$, $25\%/75\%$, $10\%/90\%$, $3.3\%/96.7\%$, or $0\%/100\%$ (the odd-ball $3.3\%$ composition corresponds to one of the $30$ images per demographic group being SD-generated with the other $29$ real).

\subsection{Healing}
\label{subsec:healing}

We define model healing as the process of retraining a generative model on only real images. The goal is to determine whether a self-poisoned model can be healed to produce images similar to the base model. This healing process proceeds in the same way as the self-poisoning described in Section~\ref{subsec:poisioning} except that the images are real, not AI generated.

\begin{figure}[t]
    \centering
    \begin{tabular}{c@{\hspace{0.1cm}}c@{\hspace{0.1cm}}c@{\hspace{0.1cm}}c@{\hspace{0.1cm}}c}
        \includegraphics[width=0.19\linewidth]{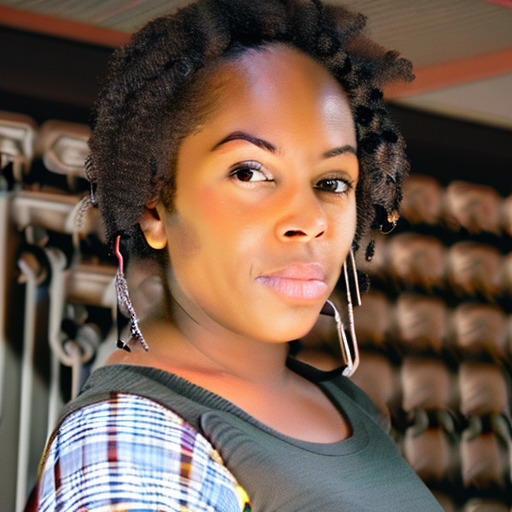} &
        \includegraphics[width=0.19\linewidth]{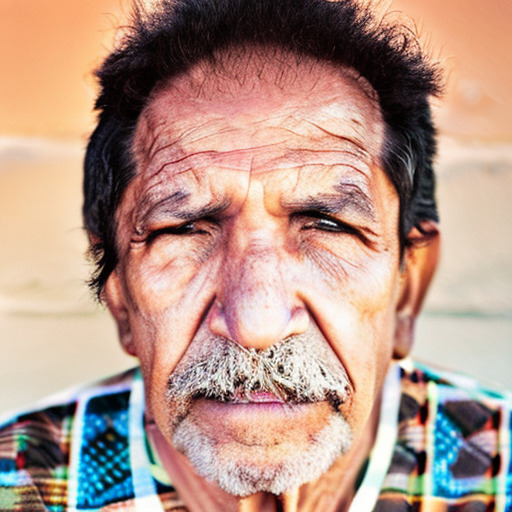} &
        \includegraphics[width=0.19\linewidth]{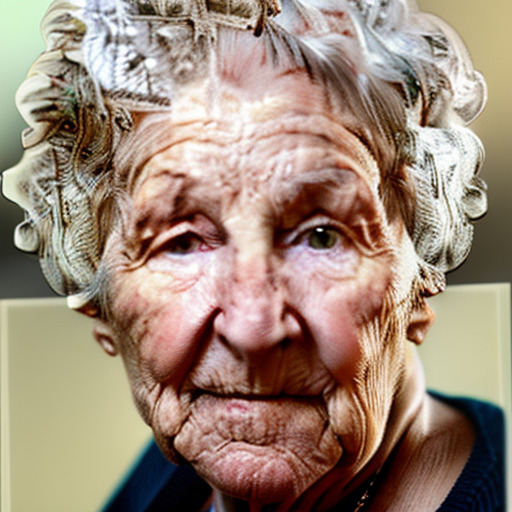} &
        \includegraphics[width=0.19\linewidth]{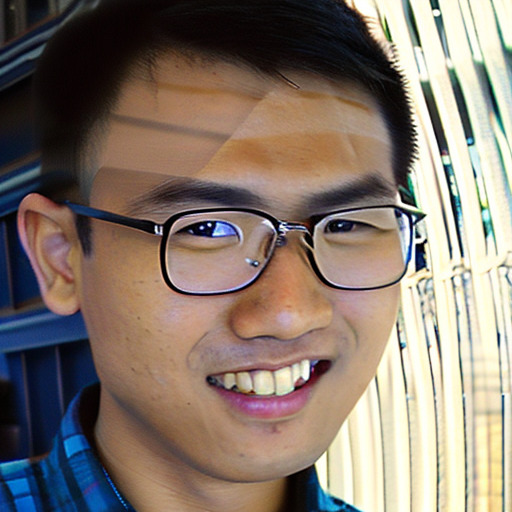} &
        \includegraphics[width=0.19\linewidth]{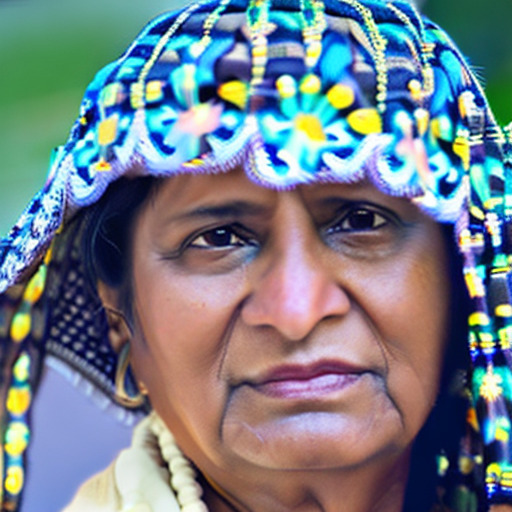}
    \end{tabular}
    \vspace{-0.25cm}
    \caption{Examples of low-quality generated images that are replaced in the retraining control experiment.}
    \label{fig:outliers}
\end{figure}
%
%

\subsection{Controls}
\label{subsec:controls}

We carried out two control experiments to determine the impact of our iterative retraining of the SD model (Section~\ref{subsec:poisioning}). Having noticed that, compared to real images, SD-generated images tend to be of higher contrast and saturation, we wondered if these color differences would impact the iterative retraining. In this first control, the color histogram of each generated image is matched to a real image. Each generated image is histogram matched -- in the three-channel luminance/chrominance (YCbCr) space -- to a real image with the most similar color distribution (measured as the image with the minimal Kullback-Leibler divergence averaged across all three YCbCr channels). This histogram matching is performed on each retraining iteration.

We also noticed that among the $900$ generated images there are occasional images with obvious artifacts, including misshapen facial features, as shown in Figure~\ref{fig:outliers}. In this second control, we removed from the retraining dataset any generated image with a single-image FID score~\citep{shaham2019singan} greater than the mean single-image FID between the first batch of $900$ generated images and their corresponding $900$ real images. This culling is performed on each retraining iteration.


%
%
\begin{figure}[t]
    \begin{center}
    \begin{tabular}{c@{\hspace{0.1cm}}c@{\hspace{0.1cm}}c@{\hspace{0.1cm}}c@{\hspace{0.1cm}}c@{\hspace{0.1cm}}c}
        \includegraphics[width=0.19\linewidth]{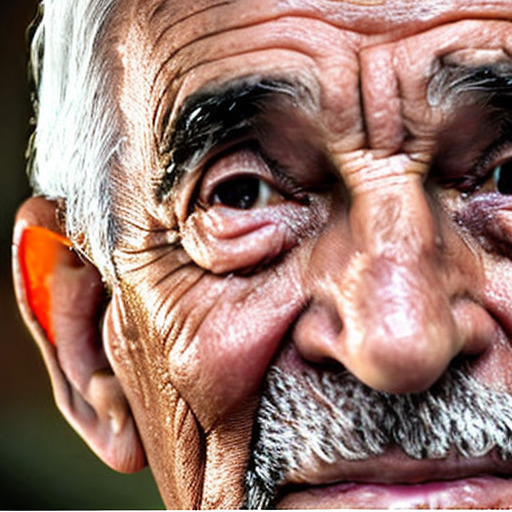} &
        \includegraphics[width=0.19\linewidth]{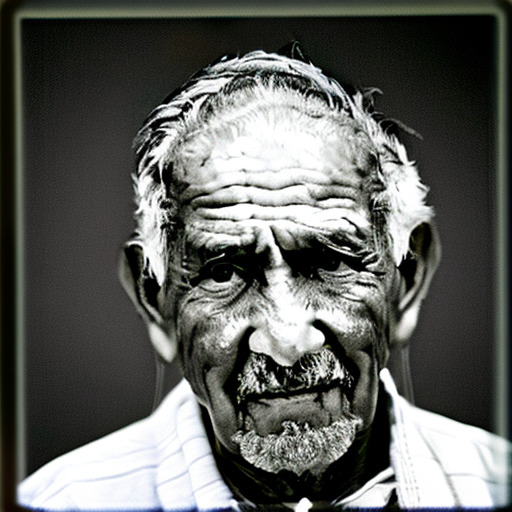} &
        \includegraphics[width=0.19\linewidth]{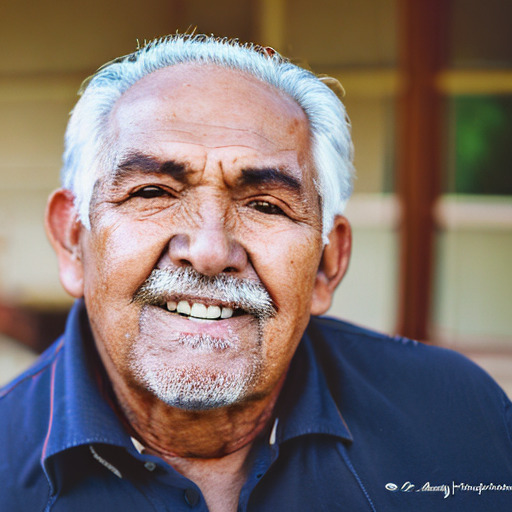} &
        \includegraphics[width=0.19\linewidth]{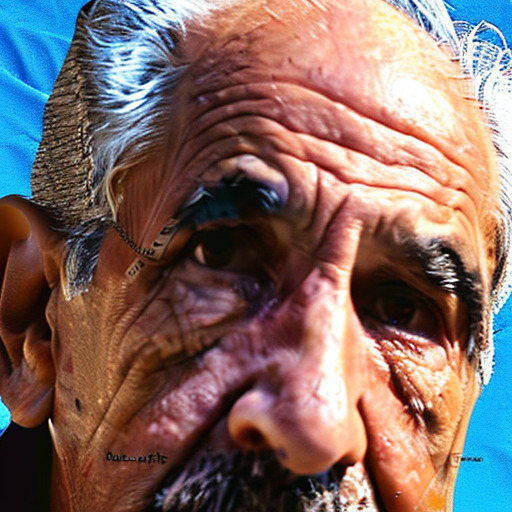} &
        \includegraphics[width=0.19\linewidth]{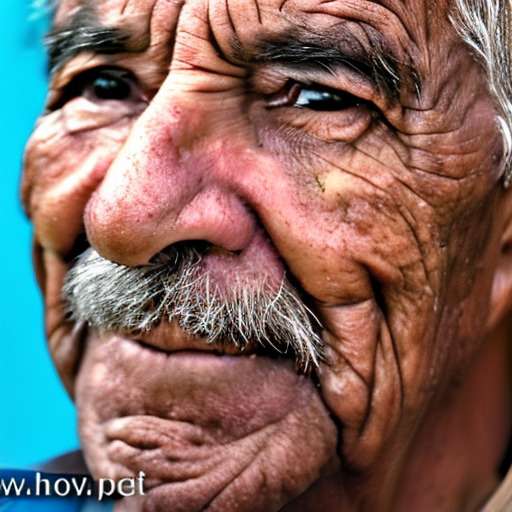} 
    \end{tabular}
    \end{center}
    \caption{Examples of images generated by the baseline version of Stable Diffusion (prompt: ``older hispanic man'').}
    \label{fig:healthy}
\end{figure}
\begin{figure}[p]
    \centering
    \begin{tabular}{r |@{\hspace{0.1cm}}c@{\hspace{0.1cm}}c@{\hspace{0.1cm}}c@{\hspace{0.1cm}}c@{\hspace{0.1cm}}c@{\hspace{0.1cm}}c}
        & \multicolumn{5}{c}{{\bf iteration}} \\
        {\bf poison} & 1 & 2 & 3 & 4 & 5  \\
        \raisebox{1cm}{$0\%$} &
        \includegraphics[width=0.15\linewidth]{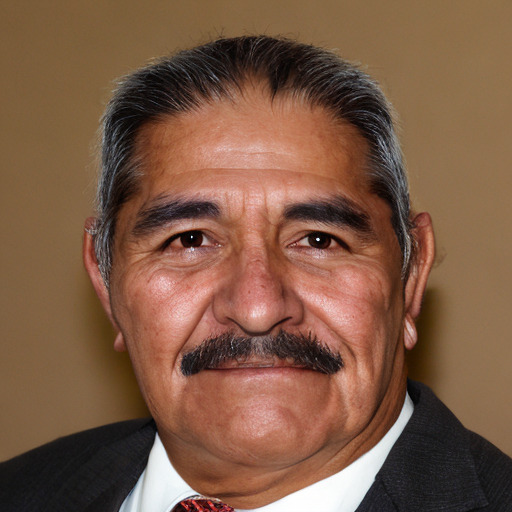} &
        \includegraphics[width=0.15\linewidth]{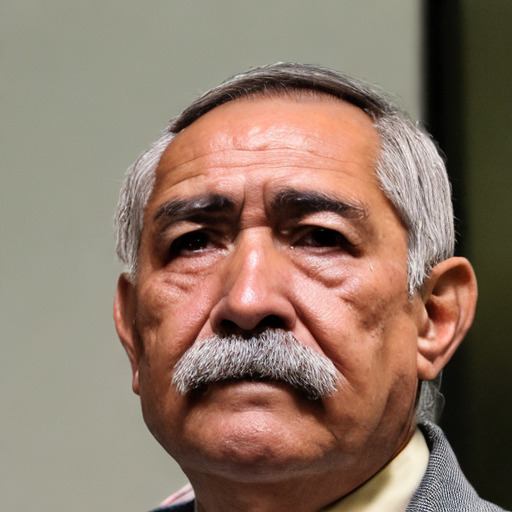} &
        \includegraphics[width=0.15\linewidth]{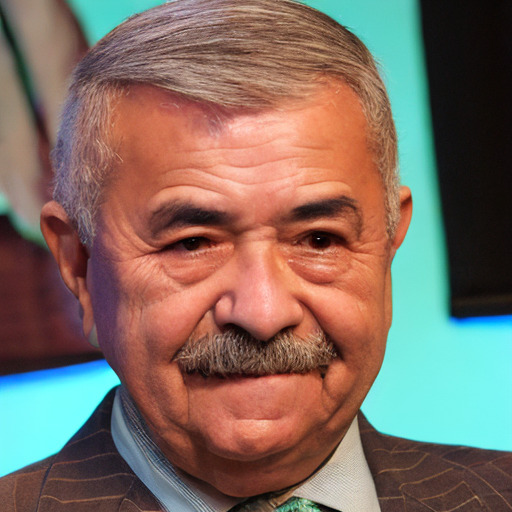} &
        \includegraphics[width=0.15\linewidth]{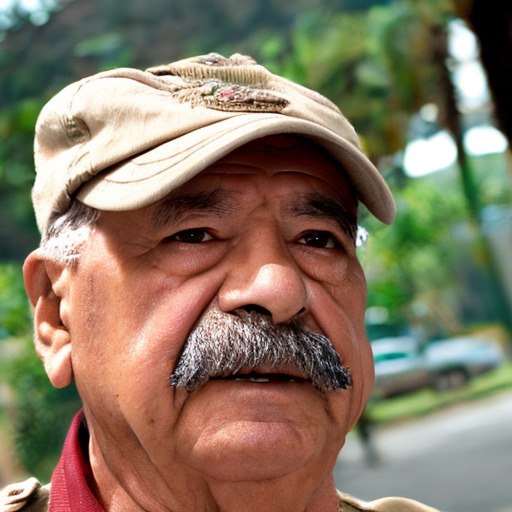} &
        \includegraphics[width=0.15\linewidth]{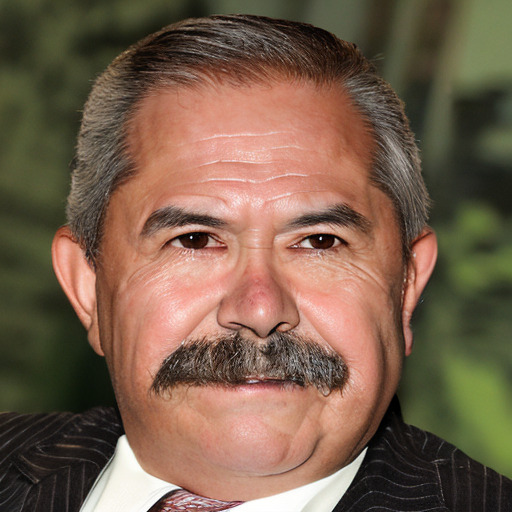} &
        \\
        \raisebox{1cm}{$3.3\%$} &
        \includegraphics[width=0.15\linewidth]{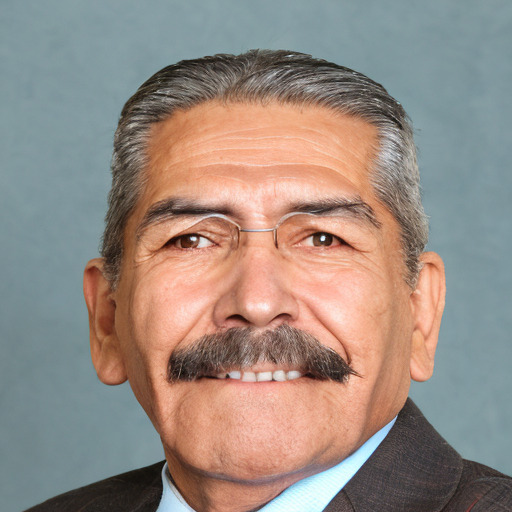} &
        \includegraphics[width=0.15\linewidth]{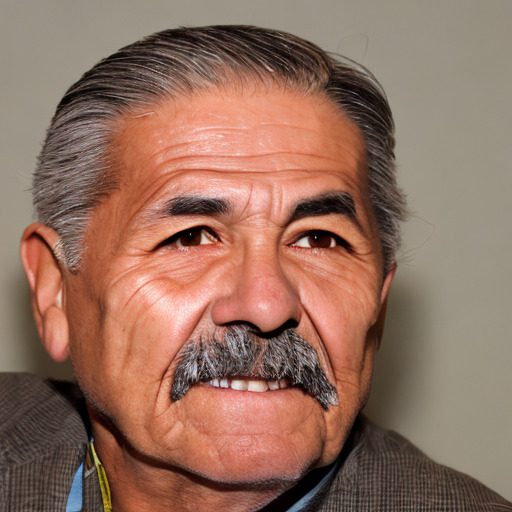} &
        \includegraphics[width=0.15\linewidth]{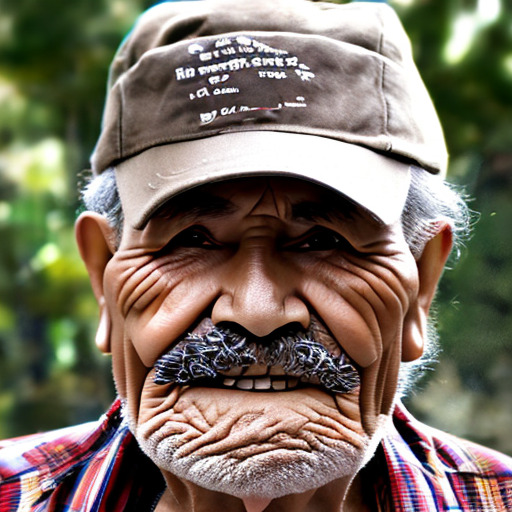} &
        \includegraphics[width=0.15\linewidth]{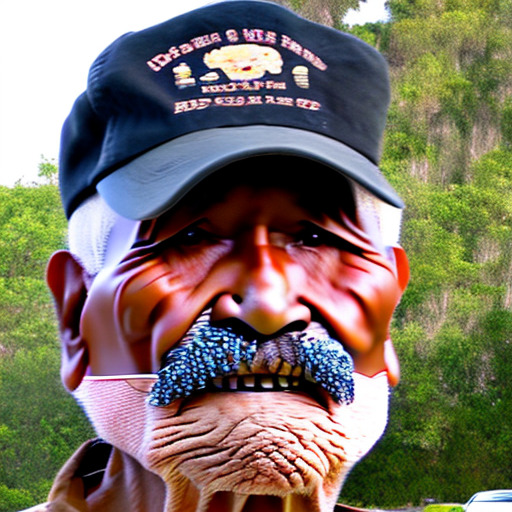} &
        \includegraphics[width=0.15\linewidth]{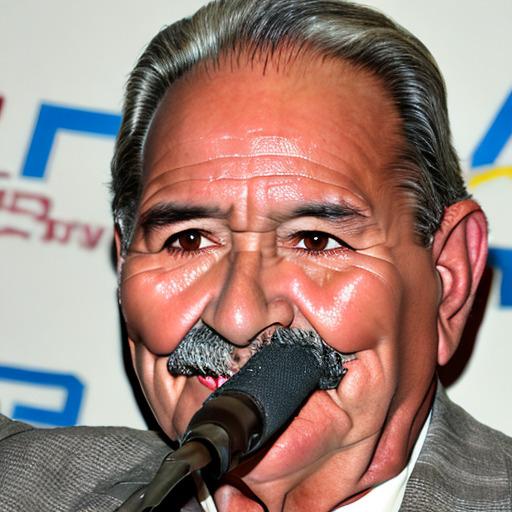}
        \\
        \raisebox{1cm}{$10\%$} &
        \includegraphics[width=0.15\linewidth]{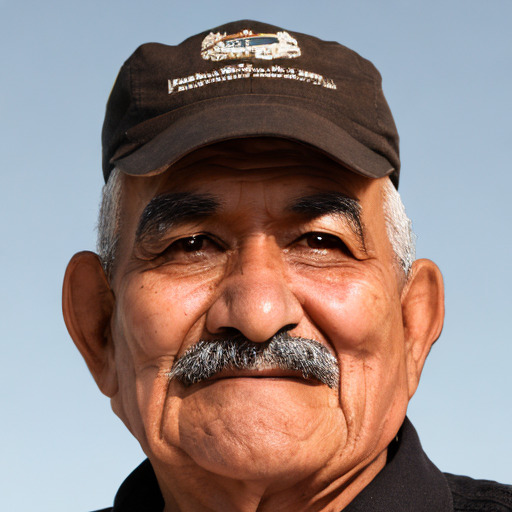} &
        \includegraphics[width=0.15\linewidth]{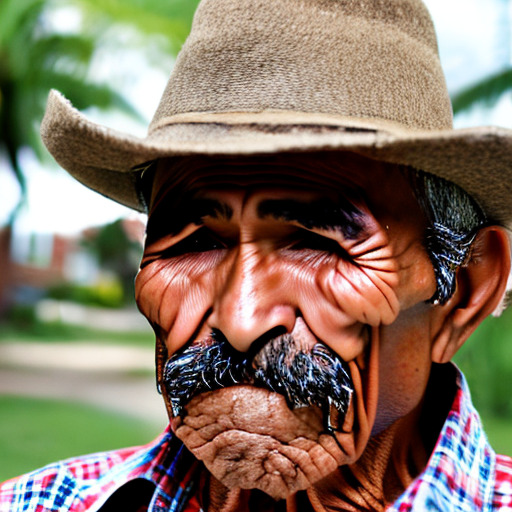} &
        \includegraphics[width=0.15\linewidth]{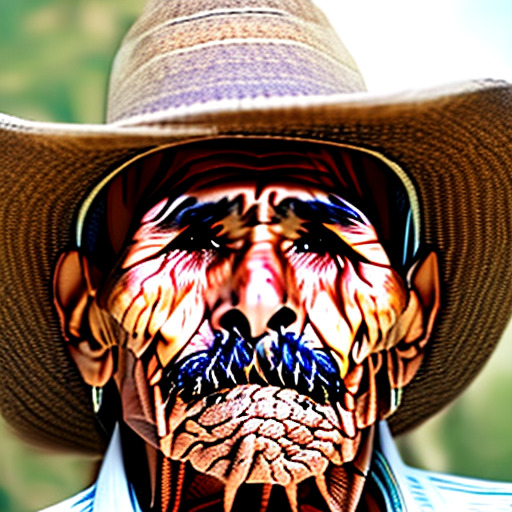} &
        \includegraphics[width=0.15\linewidth]{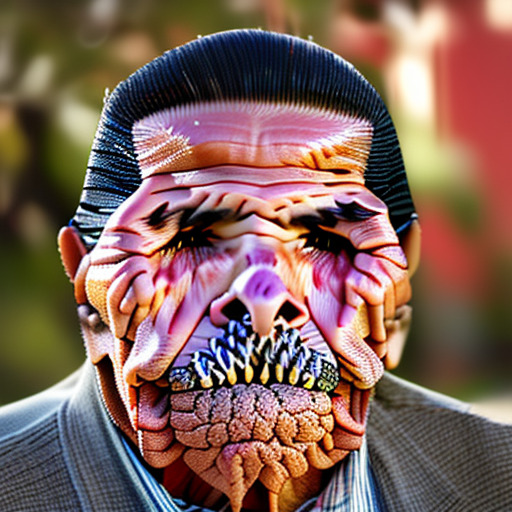} &
        \includegraphics[width=0.15\linewidth]{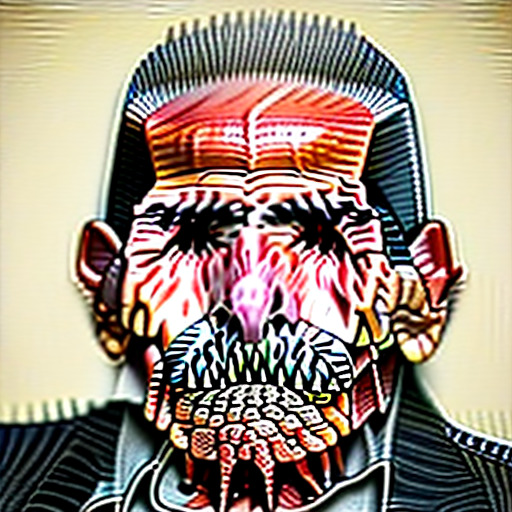} 
        \\
        \raisebox{1cm}{$25\%$} &
        \includegraphics[width=0.15\linewidth]{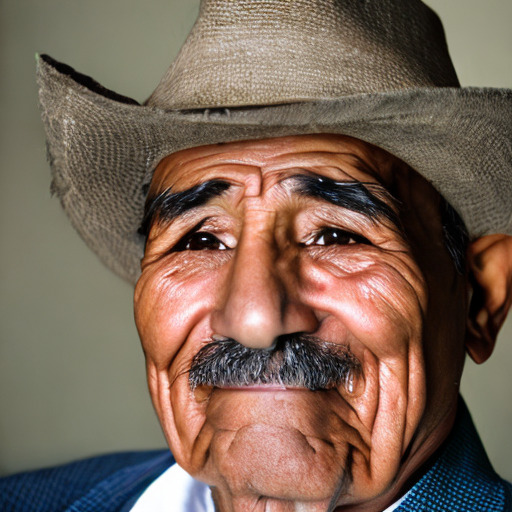} &
        \includegraphics[width=0.15\linewidth]{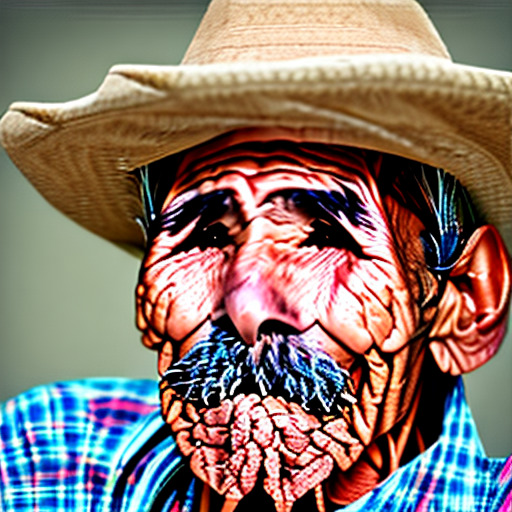} &
        \includegraphics[width=0.15\linewidth]{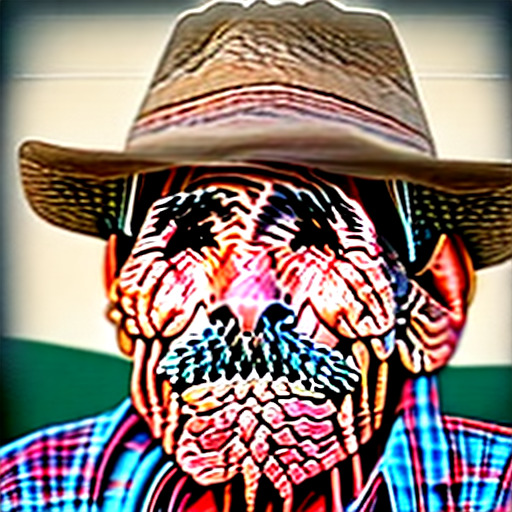} &
        \includegraphics[width=0.15\linewidth]{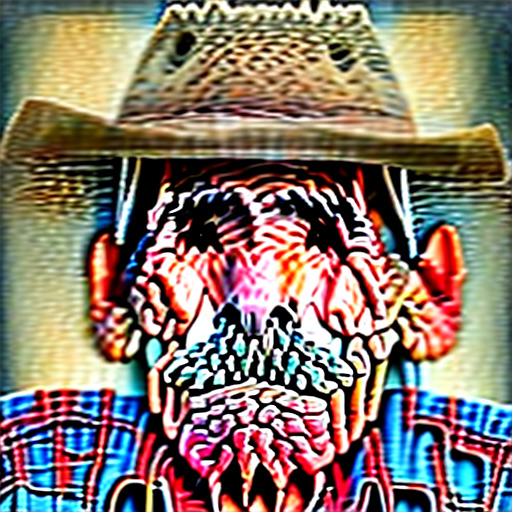} &
        \includegraphics[width=0.15\linewidth]{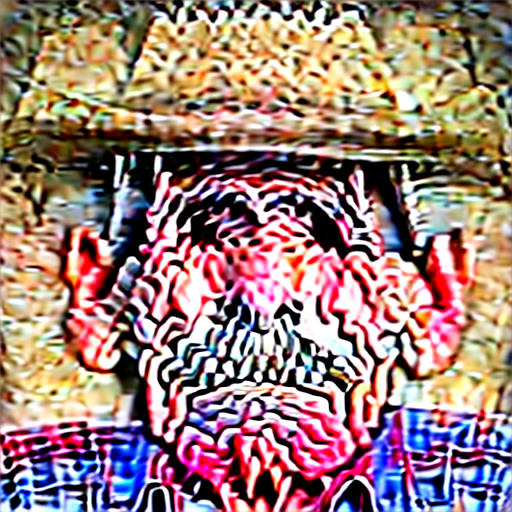} 
        \\
        \raisebox{1cm}{$50\%$} &
        \includegraphics[width=0.15\linewidth]{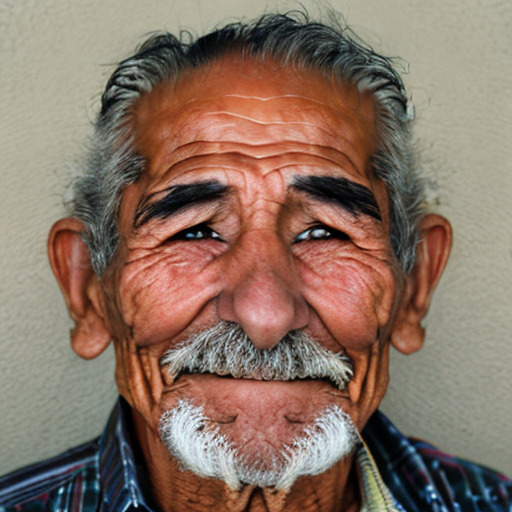} &
        \includegraphics[width=0.15\linewidth]{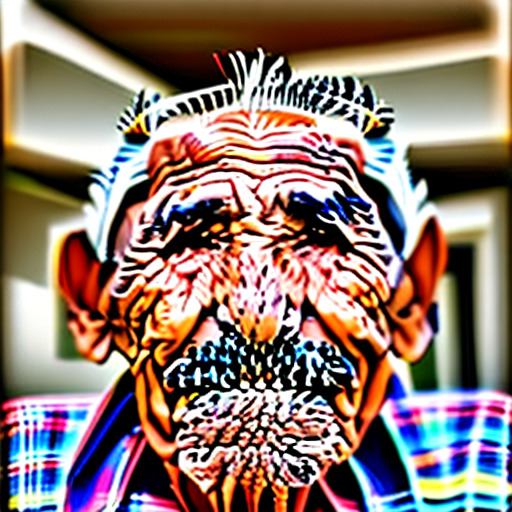} &
        \includegraphics[width=0.15\linewidth]{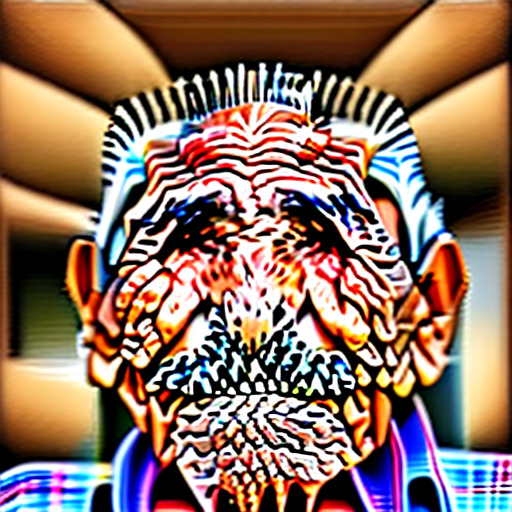} &
        \includegraphics[width=0.15\linewidth]{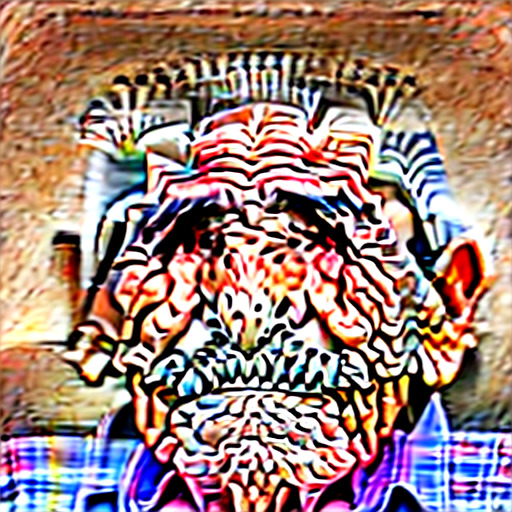} &
        \includegraphics[width=0.15\linewidth]{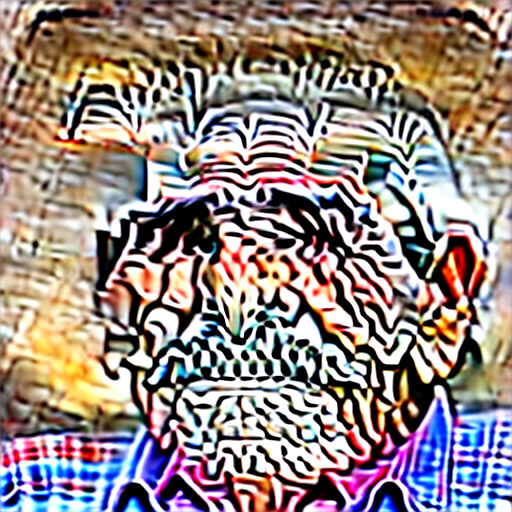} 
        \\
        \raisebox{1cm}{$100\%$} &
        \includegraphics[width=0.15\linewidth]{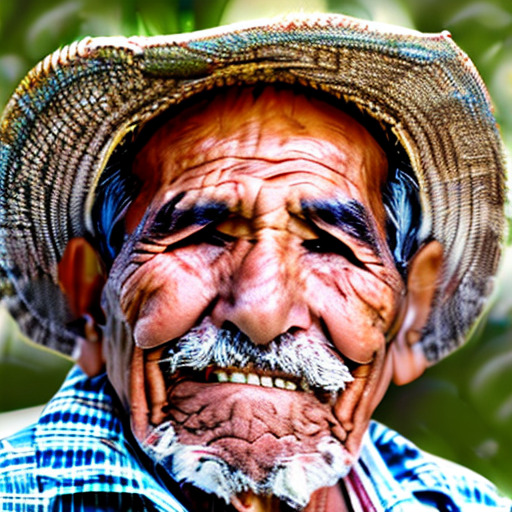} &
        \includegraphics[width=0.15\linewidth]{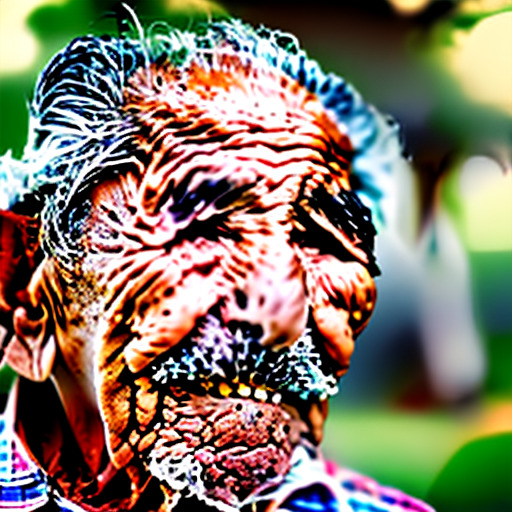} &
        \includegraphics[width=0.15\linewidth]{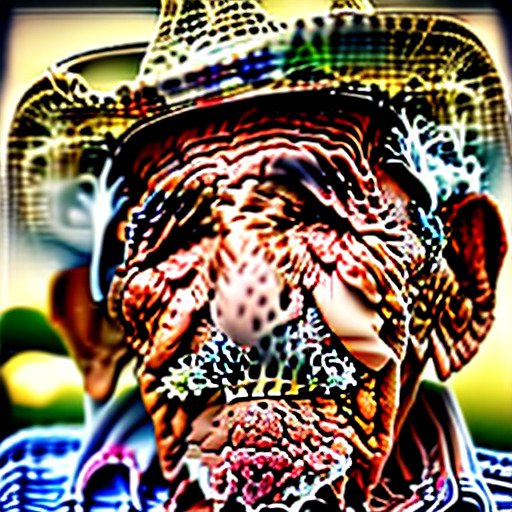} &
        \includegraphics[width=0.15\linewidth]{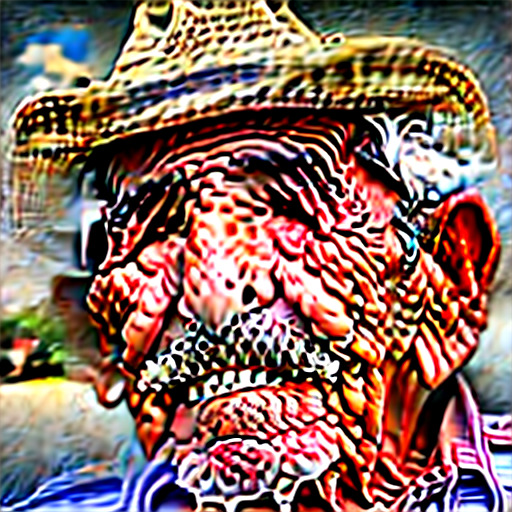} &
        \includegraphics[width=0.15\linewidth]{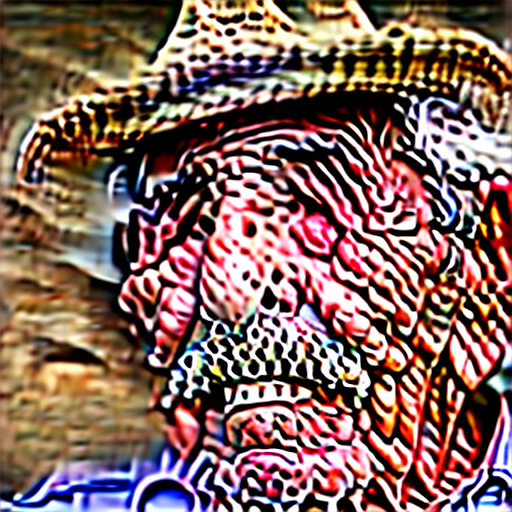} 
        \\
        \hline
        & \multicolumn{5}{c}{{\bf example}} \\
        {\bf poison} & 1 & 2 & 3 & 4 & 5  \\
        \raisebox{1cm}{$0\%$} & 
        \includegraphics[width=0.15\linewidth]{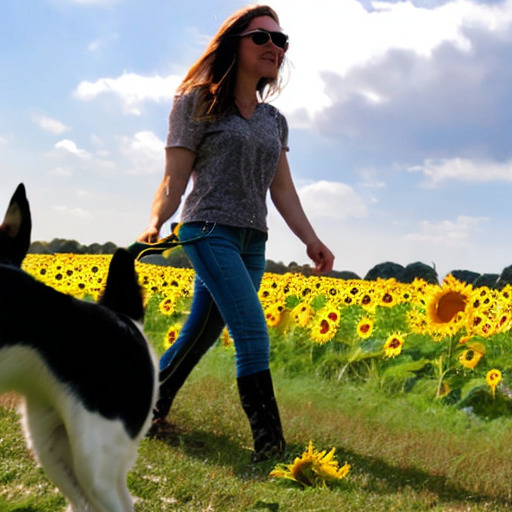} &
        \includegraphics[width=0.15\linewidth]{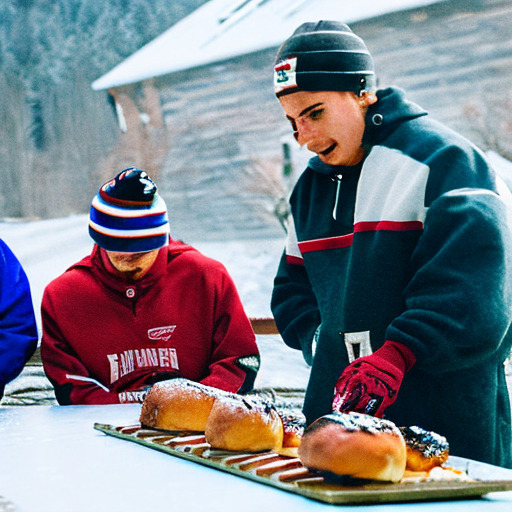} &
        \includegraphics[width=0.15\linewidth]{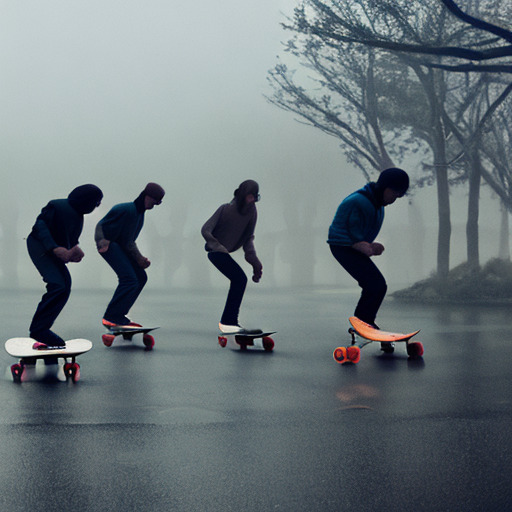} &
        \includegraphics[width=0.15\linewidth]{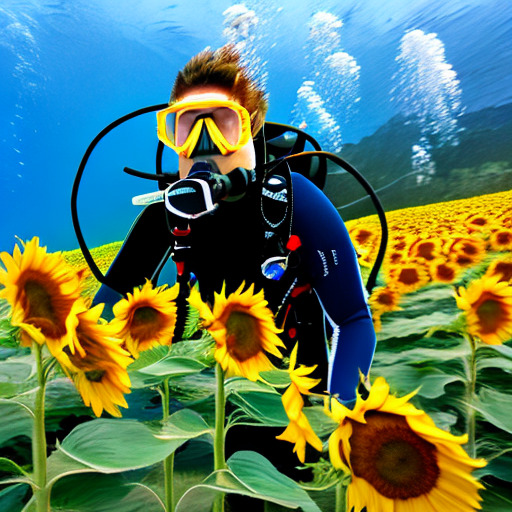} &
        \includegraphics[width=0.15\linewidth]{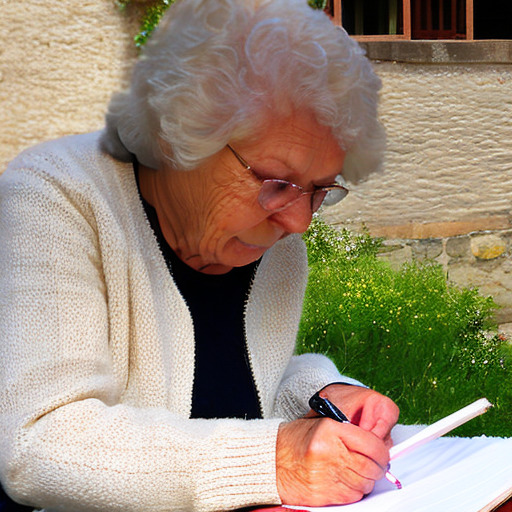} \\
        \raisebox{1cm}{$25\%$} &
        \includegraphics[width=0.15\linewidth]{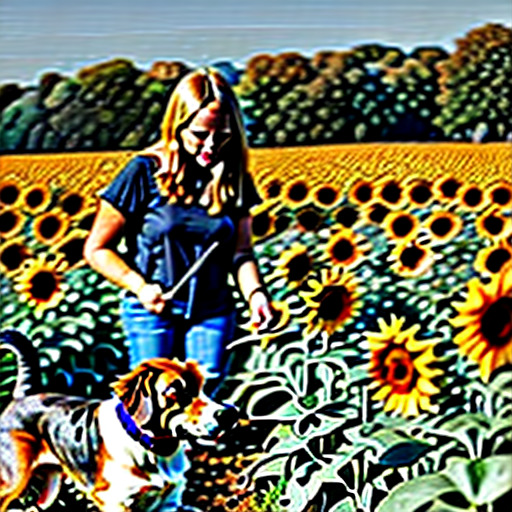} &
        \includegraphics[width=0.15\linewidth]{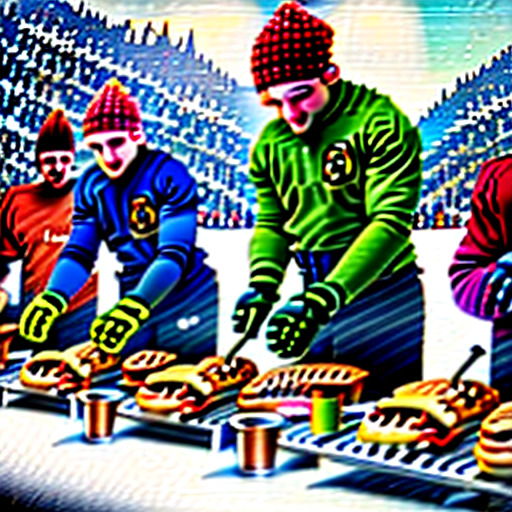} &
        \includegraphics[width=0.15\linewidth]{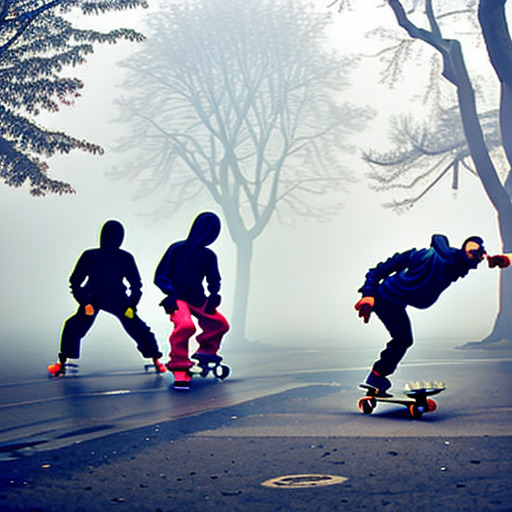} &
        \includegraphics[width=0.15\linewidth]{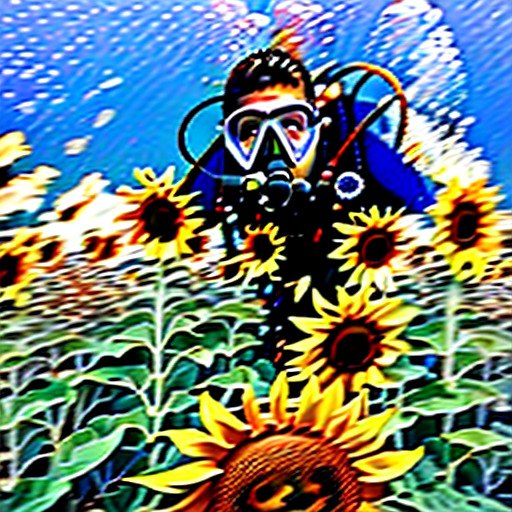} &
        \includegraphics[width=0.15\linewidth]{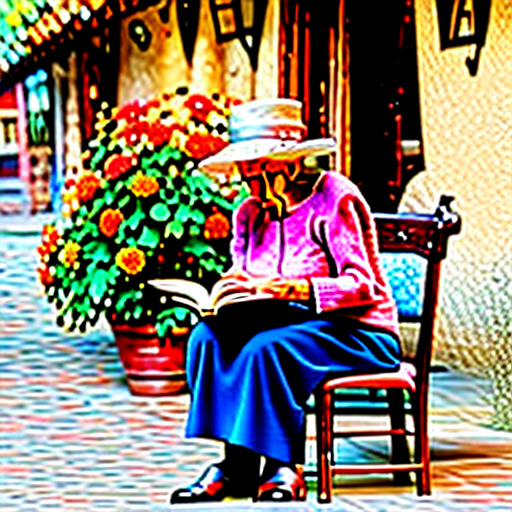}
    \end{tabular}
    \caption{Examples generated after iterative retraining for different compositions of the retraining dataset:  $0\%$ SD-generated and $100\%$ real to $100\%$ SD-generated faces and $0\%$ real. Shown in the lower panel are examples generated with text prompts distinct from those used in the retraining.}
    \label{fig:iterations}
\end{figure}

\section{Results}

Shown in Figure~\ref{fig:healthy} are five representative images generated from the baseline Stable Diffusion model for a single demographic group (``older hispanic man''). Generally speaking, images from the baseline model are consistent with the text prompt and of high visual quality.

\begin{figure}[t!]
    \centering
    \includegraphics[height=0.38\linewidth]{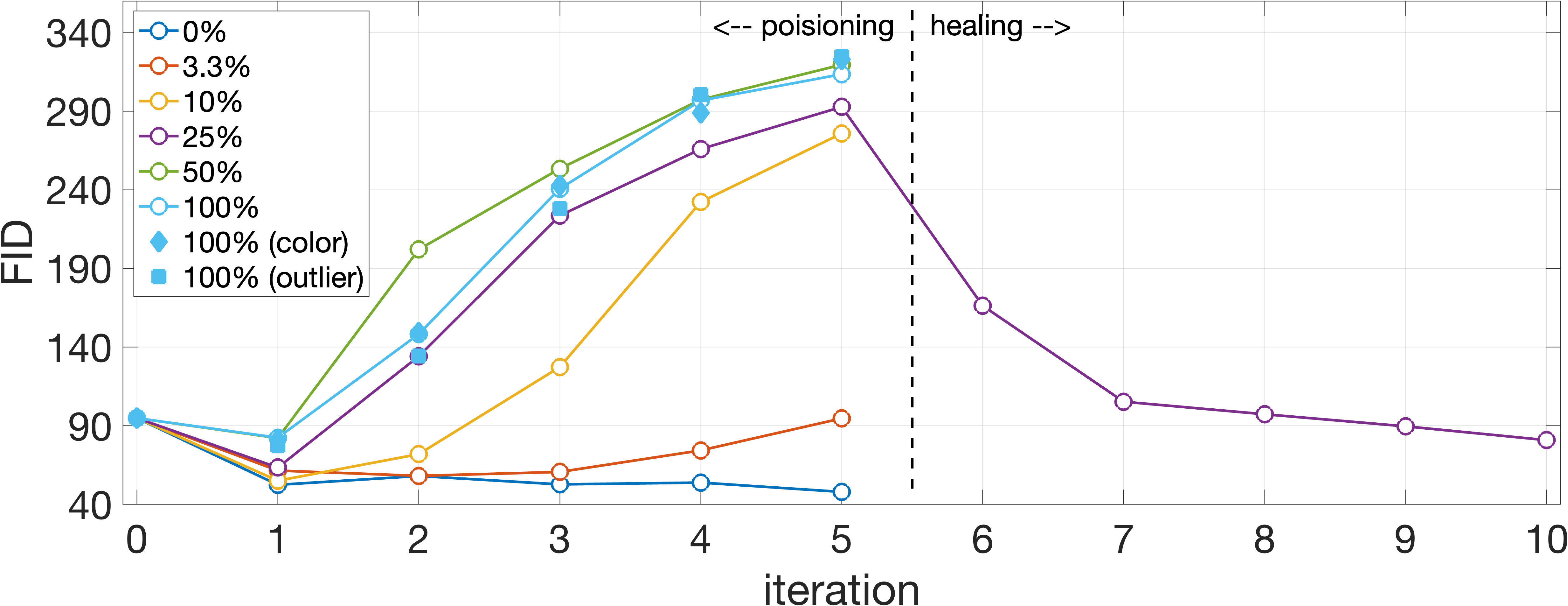} \\
    \vspace{0.5cm}
    \includegraphics[height=0.38\linewidth]{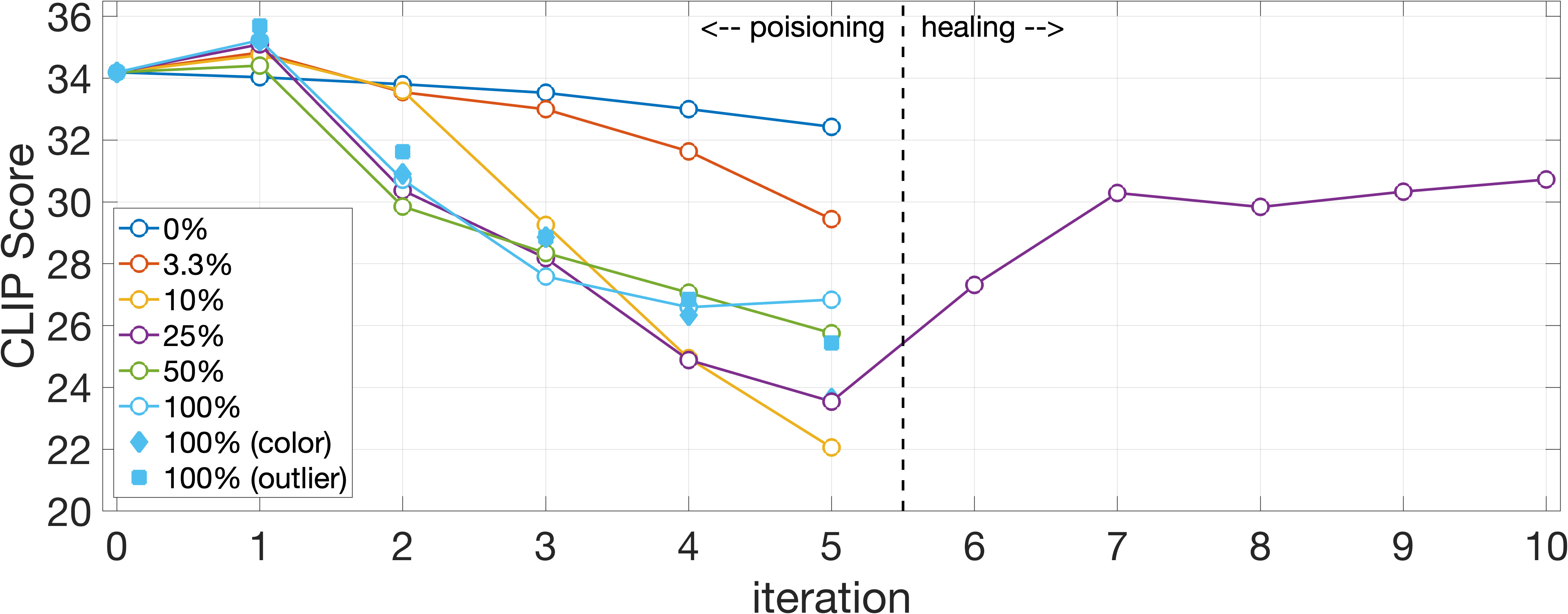}    
    \caption{Shown are the FID and CLIP as a function of the number of retraining iterations and the composition of the retraining dataset ranging from $100\%$ SD-generated faces and $0\%$ real faces to $0\%$ SD-generated and $100\%$ real (``poisoning''). The diamond plot symbol corresponds to the $100\%$/$0\%$ condition in which the retraining dataset is color matched to the real faces. The square plot symbol corresponds to the the same condition in which the retraining dataset was curated on each iteration to remove low quality faces. The trend is the same for both metrics: the presence of generated faces leads to a degradation in quality across iterations (a higher FID and a lower CLIP correspond to lower image quality). Also shown is the FID and CLIP score for the $25\%$ model retrained on an additional five iterations ($6$-$10$) on only real images (``healing'').}
    \label{fig:fid-clip}
\end{figure}

Shown in the first row of Figure~\ref{fig:iterations} are representative images generated from iterative retraining of the baseline SD model on images of real faces taken from the FFHQ facial dataset. The generated images are semantically consistent with the text prompts, exhibit the prototypical alignment property of the faces in the FFHQ dataset, and show no signs of distortion. Shown in Figure~\ref{fig:fid-clip} are the FID and CLIP scores for the full set of generated images (labeled $0\%$), from which we see that iterative retraining on real images causes no degradation in the resulting model.

Also shown in the top portion of Figure~\ref{fig:iterations} are representative images generated from iterative retraining of the baseline SD model with a different mixture of self-generated and real images. Regardless of the mixture ratio, the iterative retraining eventually leads to collapse by the fifth iteration, at which point the generated images are highly distorted. This model collapse can be seen qualitatively by the appearance of the generated images and quantitatively by the significant deviation of the FID and CLIP score from the baseline model. This collapse is apparent in Figure~\ref{fig:fid-clip} where we see that after a small improvement in quality on the first iteration, both the FID and CLIP scores reveal a significant degradation in image quality (a high FID and a low CLIP correspond to lower quality images).

The filled plot symbols (diamond and square) in Figure~\ref{fig:iterations} correspond to the two control conditions in which the retraining dataset is color matched to real images (diamond) and any low-quality generated images are replaced with high-quality generated images prior to retraining (square). Even these curated retraining datasets lead to model collapse at the same rate as the other datasets.

In addition to the degradation in image quality, and consistent with previous reports~\citep{hataya2023will}, we also note that model self-poisoning leads to a lack of diversity in terms of the appearance of the generated faces. This can be seen in Figure~\ref{fig:iterations} where, particularly when the self-poisoning is greater than $10\%$, the generated faces are highly similar across the latter iterations.

Shown in the lower two rows of Figure~\ref{fig:iterations} are representative examples of images generated with text prompts distinct from the demographic prompts used in the model retraining. The text prompts used to generate the images in the lower panel of Figure~\ref{fig:iterations} are: ``A dog walker training a dog amidst a field of sunflowers''; ``A football team grilling hamburgers at a snowy ski resort''; ``A group of skateboarders practicing martial arts in a mysterious foggy landscape''; ``A marine biologist scuba diving amidst a field of sunflowers''; and ``An elderly woman writing in a journal in a charming village square''. The images generated by the model retrained on entirely real images ($0\%$) produce semantically coherent images with no obvious visual artifacts. The images generated by the model retrained on $25\%$ SD-generated faces often exhibit the same textured artifacts as seen in the faces in the upper portion of this figure. This means that the model self-poisoning is not limited to a specific category of images used in the retraining but seems to impact the model more broadly.

\begin{figure}[t!]
    \centering
    \begin{tabular}{@{\hspace{0.1cm}}c@{\hspace{0.1cm}}c@{\hspace{0.1cm}}c@{\hspace{0.1cm}}c@{\hspace{0.1cm}}c@{\hspace{0.1cm}}c}
        \multicolumn{5}{c}{{\bf iteration}} \\
        6 & 7 & 8 & 9 & 10  \\
        \includegraphics[width=0.19\linewidth]{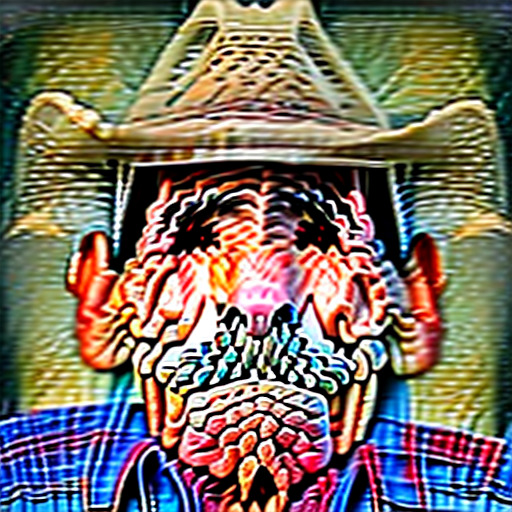} &
        \includegraphics[width=0.19\linewidth]{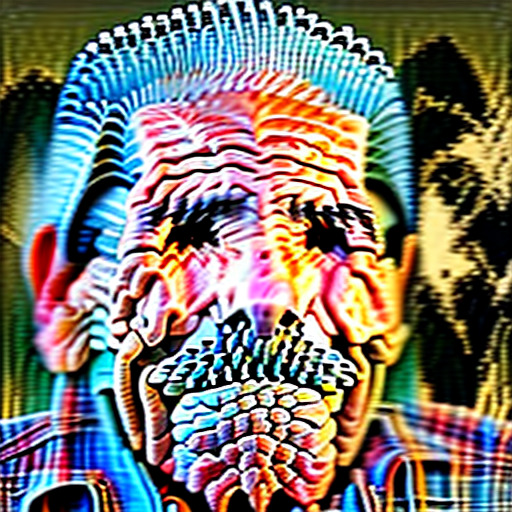} &
        \includegraphics[width=0.19\linewidth]{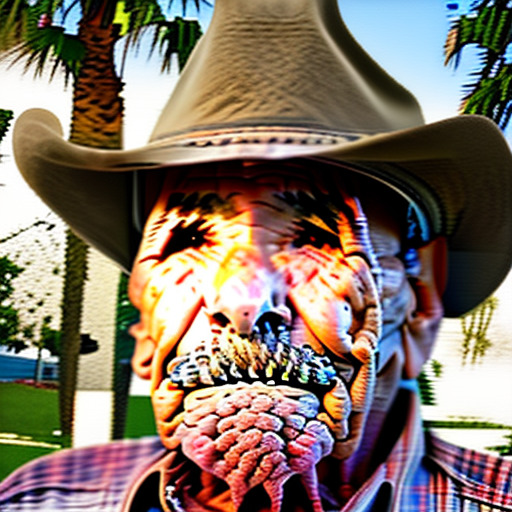} &
        \includegraphics[width=0.19\linewidth]{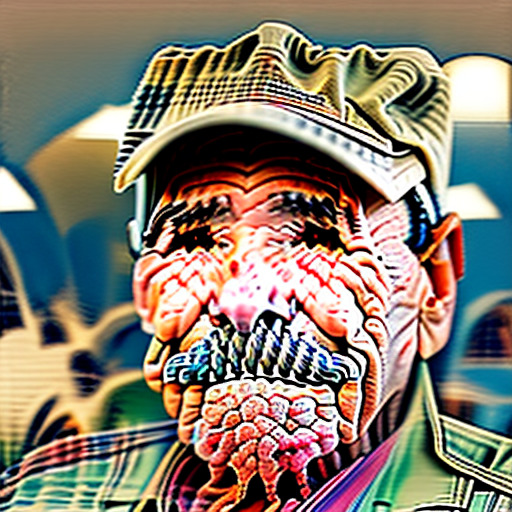} &
        \includegraphics[width=0.19\linewidth]{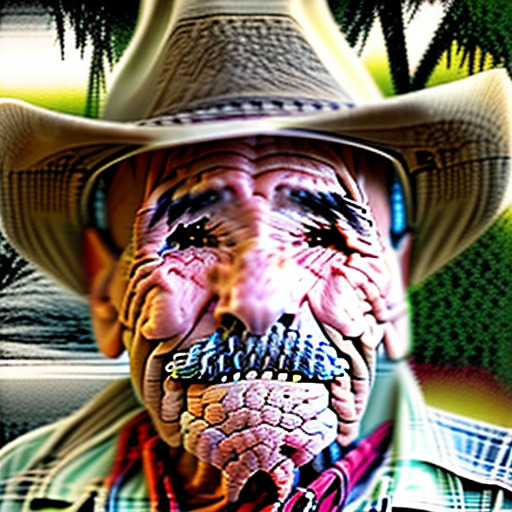}
        \\
        \includegraphics[width=0.19\linewidth]{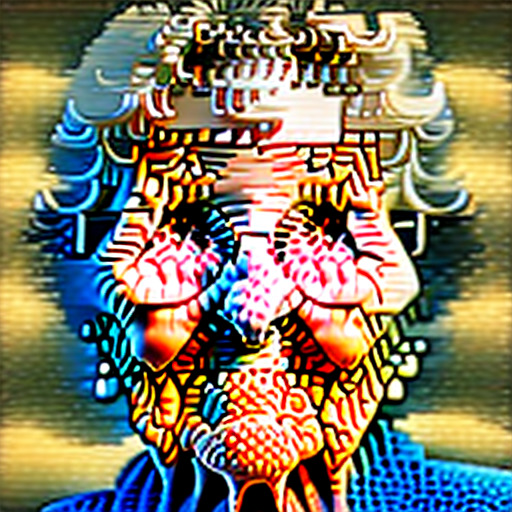} &
        \includegraphics[width=0.19\linewidth]{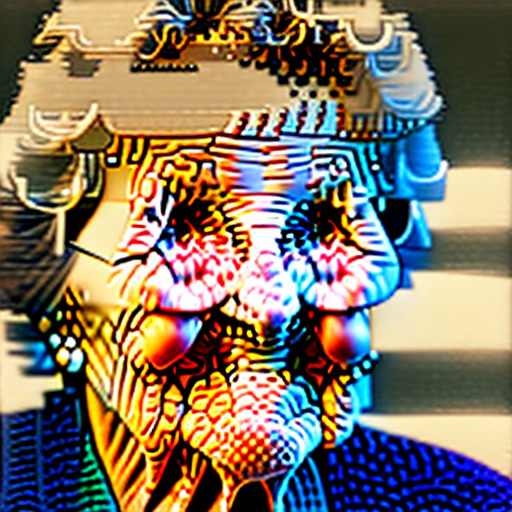} &
        \includegraphics[width=0.19\linewidth]{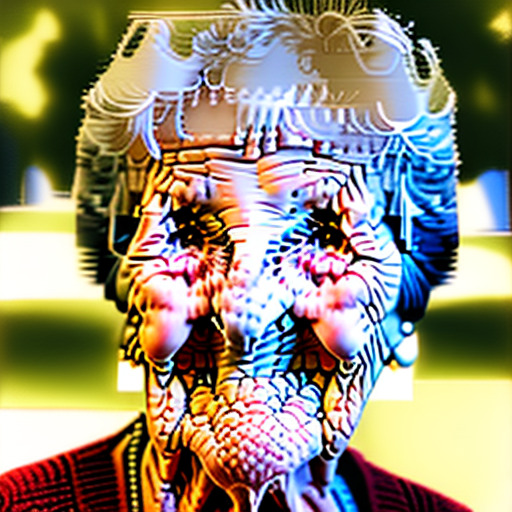} &
        \includegraphics[width=0.19\linewidth]{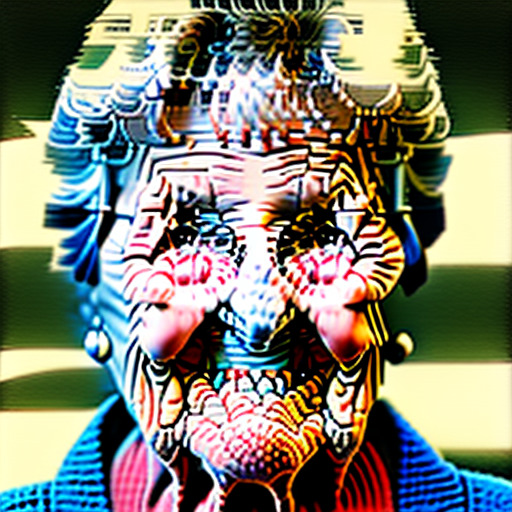} &
        \includegraphics[width=0.19\linewidth]{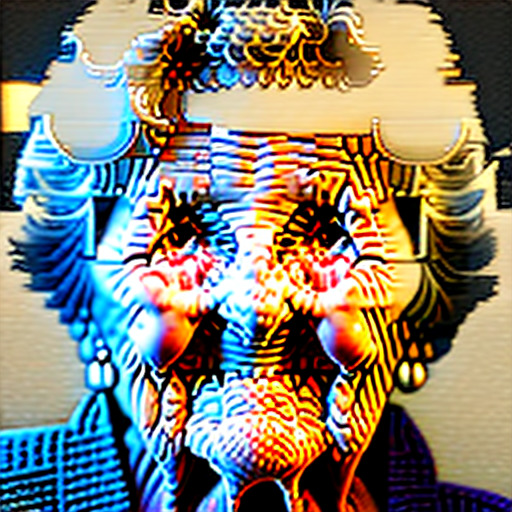} 
        \\
        \includegraphics[width=0.19\linewidth]{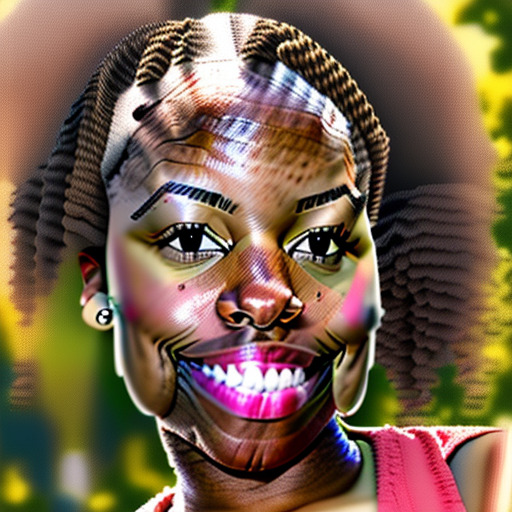} &
        \includegraphics[width=0.19\linewidth]{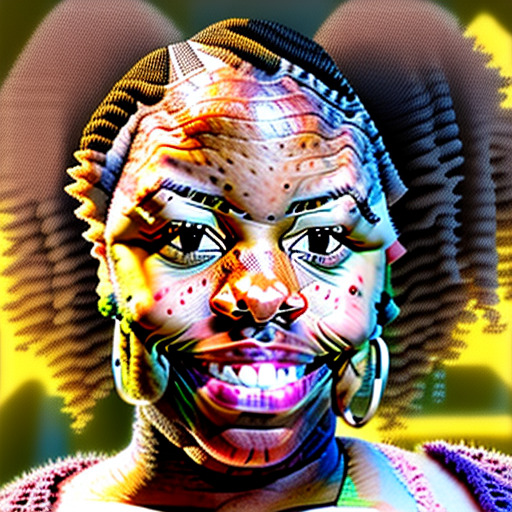} &
        \includegraphics[width=0.19\linewidth]{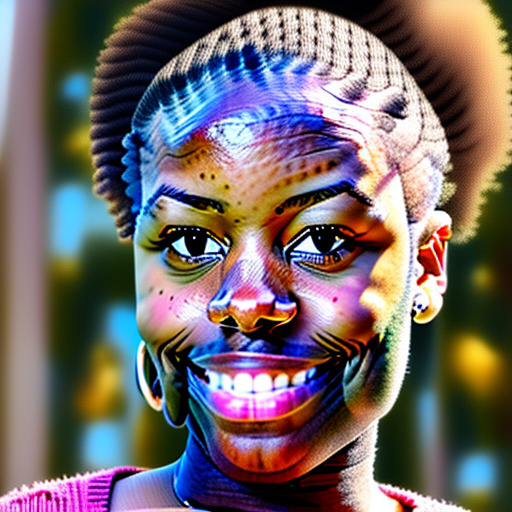} &
        \includegraphics[width=0.19\linewidth]{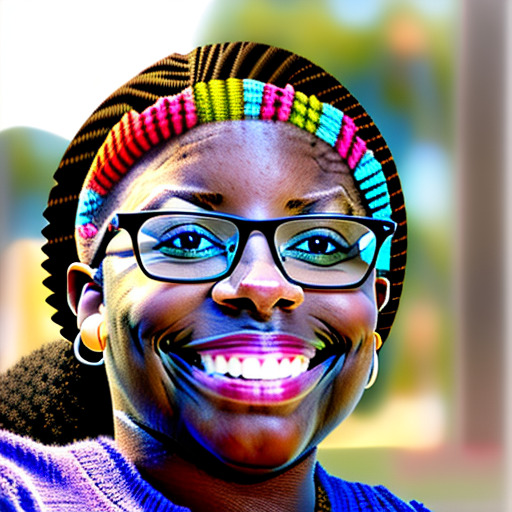} &
        \includegraphics[width=0.19\linewidth]{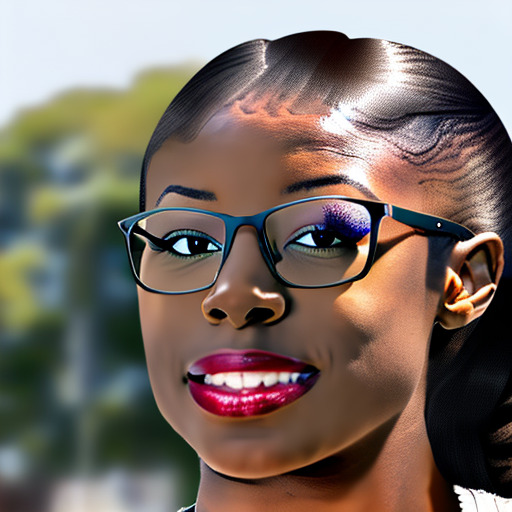}
        \\
        
        \includegraphics[width=0.19\linewidth]{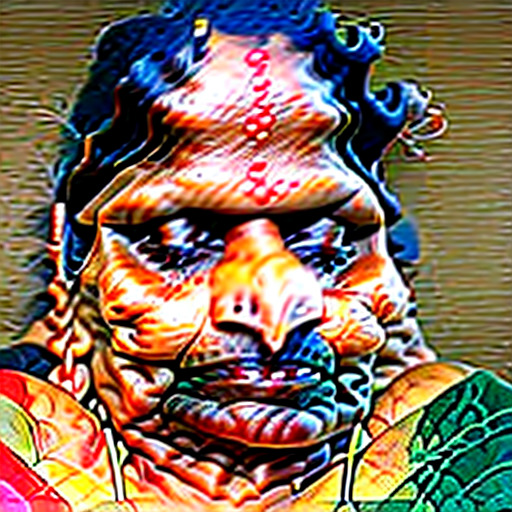} &
        \includegraphics[width=0.19\linewidth]{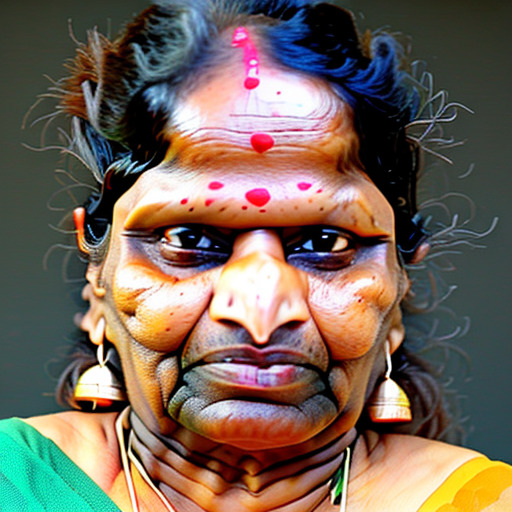} &
        \includegraphics[width=0.19\linewidth]{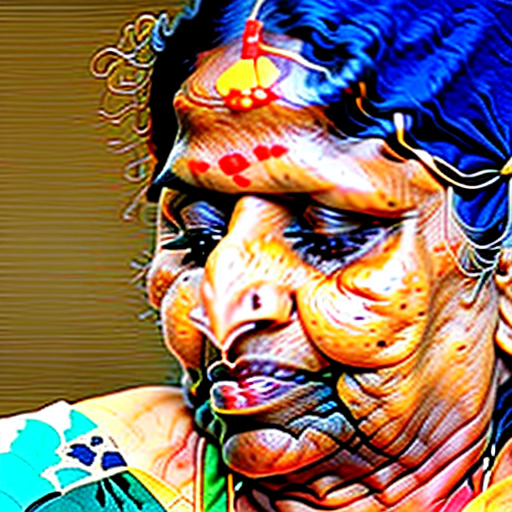} &
        \includegraphics[width=0.19\linewidth]{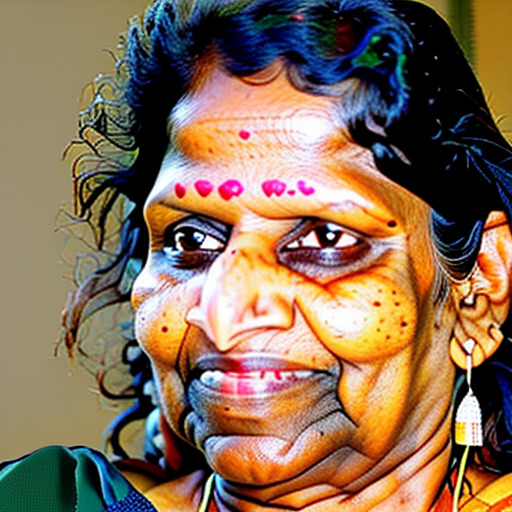} &
        \includegraphics[width=0.19\linewidth]{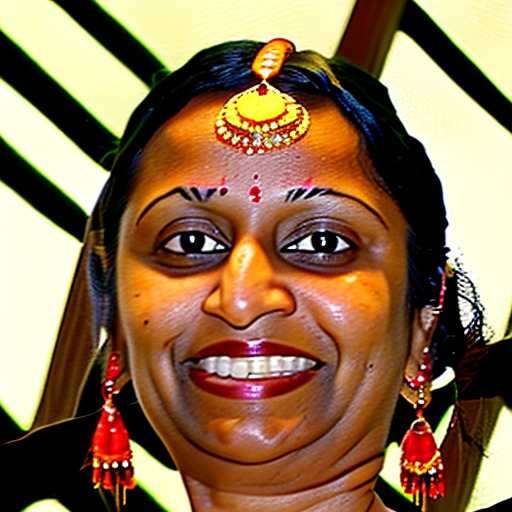}
        \\
        
        \includegraphics[width=0.19\linewidth]{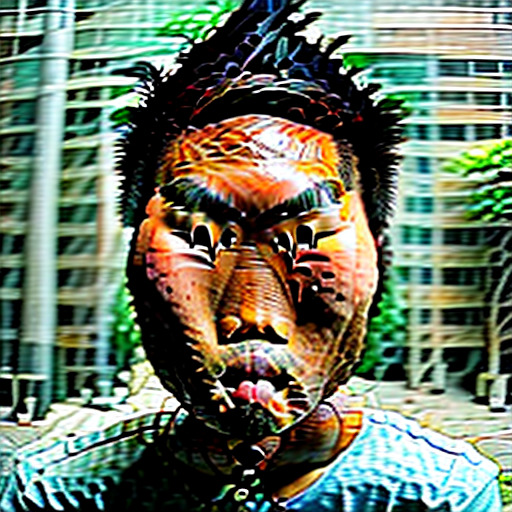} &
        \includegraphics[width=0.19\linewidth]{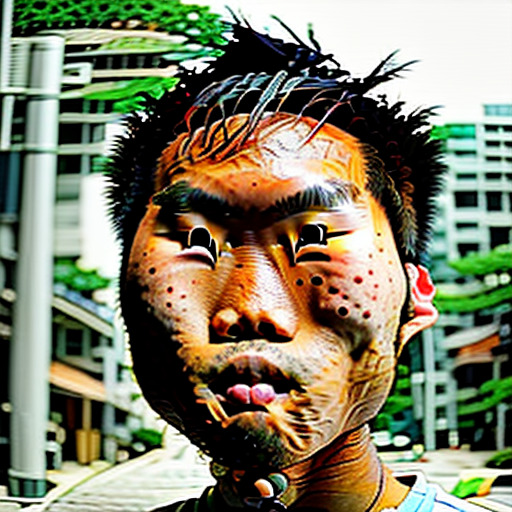} &
        \includegraphics[width=0.19\linewidth]{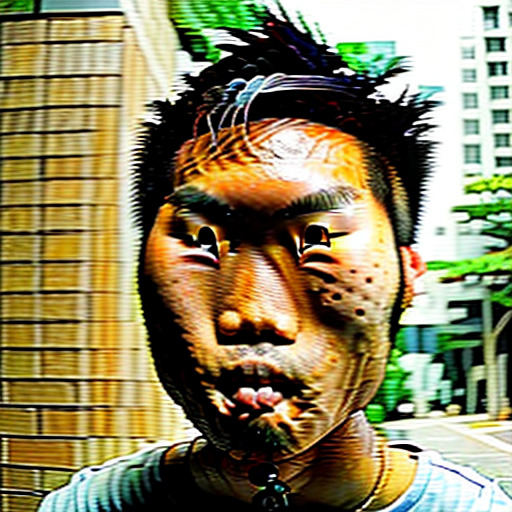} &
        \includegraphics[width=0.19\linewidth]{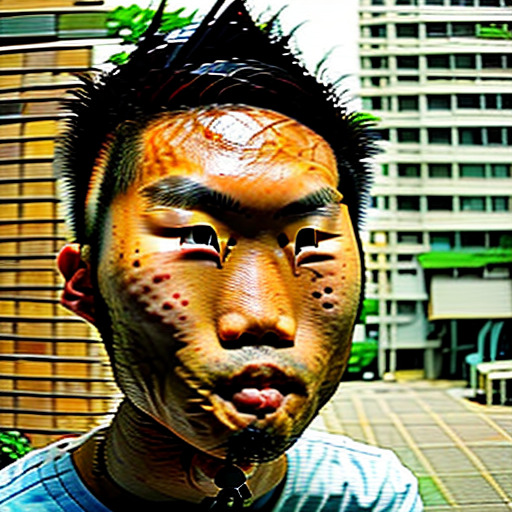} &
        \includegraphics[width=0.19\linewidth]{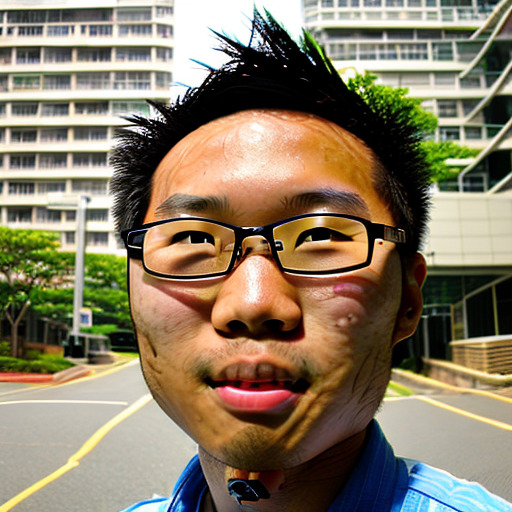}
        
       \end{tabular}
    \caption{Examples generated after retraining of the $25\%$ self-poisoned model with only real images. In some cases, the self-poisoning persists, while in others the model appears to partially heal itself.}
    \label{fig:healing}
\end{figure}

Lastly, we wondered if, once self-poisoned, the model could be ``healed'' by retraining on only real images.. The model self-poisoned for five iterations with $25\%$ SD-generated images was retrained for another five iterations on only real images. Shown in Figure~\ref{fig:healing} are representative examples of faces generated from five different demographic groups across these five additional iterations. Although in some cases, by the last iteration, the generated faces have fewer artifacts, in other cases, the artifacts persist. Shown in the right portion of Figure~\ref{fig:fid-clip} are the FID and CLIP scores for these healing iterations in which we see that the FID recovers to the original base model and the CLIP score almost recovers to base model levels. 

Although the mean FID and CLIP score recovers, we clearly see remnants of the self-poisoning in some of the faces in Figure~\ref{fig:healing}. This larger variation is evident in the standard deviation of the CLIP score, which is $2.8$ for the base model (with a mean of $35.1$) but is $4.2$ for the healed model after five iterations (with a mean of $27.3$). It appears that the model can partially -- but not entirely -- heal.
        

\section{Discussion}

We find that at least one popular diffusion-based, text-to-image generative-AI system is surprisingly vulnerable to data poisoning with its own creations. This data poisoning can occur unintentionally by, for example, indiscriminately scraping and ingesting images from online sources. Or, it can occur from an adversarial attack where websites are intentionally populated with poisoned data, as described in~\citep{carlini2023poisoning}. Even more aggressive adversarial attacks can be launched by manipulating both the image data and text prompt on as little as $0.01\%$ to $0.0001\%$ of the dataset~\citep{carlini2021poisoning}.

In the face of these vulnerabilities, some reasonable measures could be taken to mitigate these risks. First, there is a large body of literature for classifying images as real or AI-generated (e.g.,~\citet{zhang2019detecting,chai2020makes,gragnaniello2021gan,liu2022detecting,corvi2023detection}). An ensemble of these types of detectors could be deployed to exclude AI-generated images from being ingested into a model's retraining. A complementary approach can automatically and robustly watermark all content produced by a model. This can be done after an image is generated using standard techniques~\citep{cox1999review} or can be baked into the synthesis by watermarking all the training data~\citep{yu2021artificial}. Lastly, more care can be taken to ensure the provenance of training images by, for example, licensing images from trusted sources.

These interventions are, of course, not perfect. Passive detection of AI-generated images is not full proof: a sophisticated adversary can remove a watermark, and provenance is not always available or completely reliable. Combined, however, these strategies will most likely mitigate some of the risk of data poisoning by significantly reducing the number of undesired images.

Open questions remain. What about the model or training data causes the data poisoning? Will data poisoning generalize across synthesis engines? Will, for example, Stable Diffusion retrained on DALL-E or Midjourney images exhibit the same type of model collapse? Can generative-AI systems be trained or modified to be resilient to this type of data poisoning? If it turns out to be difficult to prevent data poisoning, are there specific techniques or data sets that can accelerate healing?

\bibliography{iclr2025_conference}
\bibliographystyle{iclr2025_conference}

\end{document}